\documentclass[runningheads]{llncs}
\usepackage[T1]{fontenc}
\usepackage{graphicx}
\usepackage{amsmath}

\begin{document}

\title{MultiViPerFrOG: A Globally Optimized Multi-Viewpoint Perception Framework for Camera Motion and Tissue Deformation}

\titlerunning{MultiViPerFrOG for Estimating Camera Motion and Tissue Deformation}
%


\author{Guido Caccianiga\inst{1,2}\orcidID{0000-0002-1272-1089} \and
Julian Nubert\inst{1,2} \and
Cesar Cadena\inst{2} \and \\ Marco Hutter\inst{2} \and Katherine J. Kuchenbecker\inst{1}\orcidID{0000-0002-5004-0313}}

\authorrunning{G. Caccianiga et al.}
%
\institute{Haptic Intelligence Department, Max Planck Institute for Intelligent Systems, Stuttgart, Germany \\ \email{caccianiga;nubert;kjk@is.mpg.de} \and
Robotic Systems Lab, ETH Zurich, Switzerland}

\maketitle              

\begin{abstract}
Reconstructing the 3D shape of a deformable environment from the information captured by a moving depth camera is highly relevant to surgery. The underlying challenge is the fact that simultaneously estimating camera motion and tissue deformation in a fully deformable scene is an ill-posed problem, especially from a single arbitrarily moving viewpoint. Current solutions are often organ-specific and lack the robustness required to handle large deformations. Here we propose a multi-viewpoint global optimization framework that can flexibly integrate the output of low-level perception modules (data association, depth, and relative scene flow) with kinematic and scene-modeling priors to jointly estimate multiple camera motions and absolute scene flow. We use simulated noisy data to show three practical examples that successfully constrain the convergence to a unique solution. 
Overall, our method shows robustness to combined noisy input measures and can process hundreds of points in a few milliseconds. MultiViPerFrOG builds a generalized learning-free scaffolding for spatio-temporal encoding that can unlock advanced surgical scene representations and will facilitate the development of the computer-assisted-surgery technologies of the future.

\end{abstract}

\section{Introduction}

Recent trends in surgical computer vision aim to synergistically implement low-level perception modules (e.g., feature description, matching, depth, segmentation, tissue tracking) to beneficially shift the computational bottleneck to the higher perception layers~\cite{lin2023semantic,psychogyios2022msdesis,schmidt2023sendd}. Indeed, the outputs of these modules require further integration to be deployed for downstream tasks such as 3D scene reconstruction, visualization, and understanding. Attempts to deploy classic algorithms like simultaneous localization and mapping (SLAM) or structure from motion (SfM), usually at the core of this spatio-temporal integration process, struggle to model deformable environments, with substantial tradeoffs requiring compromises to generalization or real-time performance~\cite{schmidt2024tracking}. Though promising, recent advancements in neural computer graphics like deformable NeRFs and 4D Gaussians require precise knowledge of the camera poses and have shown limited performance in surgery when they lack multi-viewpoint input data~\cite{zhu2024endogsdeformableendoscopictissues}. 

The underlying challenge affecting all these methods is the fact that simultaneously estimating camera motion and tissue deformation in a fully deformable scene is an ill-posed problem, especially from a single arbitrarily moving viewpoint. Here, we propose a \textbf{multi-viewpoint global optimization framework that can flexibly integrate the output of low-level perception modules (data association, depth, and relative scene flow) with kinematic and scene-modeling priors to jointly estimate multiple camera motions and absolute scene flow.} By integrating the redundancy of multiple 3D vectors and kinematic measurements into a fast large-scale global optimization, we constrain the ambiguity of concurrent pose estimation and absolute deformable tissue tracking. The proposed method (MultiViPerFrOG) builds a generalized learning-free scaffolding for spatio-temporal encoding that can unlock advanced surgical scene representations. Fast free-viewpoint visualization for multi-user interfaces and partial or shared autonomy are the natural next steps in the perception pyramid for computer-integrated surgery.


\section{Background and Related Work}
Reconstructing the 3D shape of a deformable environment from the information captured by a moving camera is a photogrammetry problem that fits well in the context of endoscopy and laparoscopy~\cite{maier2013optical}. DynamicFusion~\cite{newcombe2015dynamicfusion} famously solved deformable reconstruction by gradually building a canonical model; its formulation allows the camera motion to be factored out prior to optimization via dense ICP. Despite its effectiveness, a canonical model can hardly be deployed in surgical applications involving large morphological changes. Bartoli et al.\ made substantial contributions in laparoscopy using shape-from-template~\cite{bartoli2015shape}, but results are often organ-specific and lack the robustness required to handle large deformations. Similarly, endoscopic deformable SLAM and SfM~\cite{lamarca2020defslam,recasens2021endo,rodriguez2022tracking} easily become unstable with complex 3D surfaces and long sequences. The Drunkard's Odometry~\cite{recasens2023drunkard} separates camera motion from scene-flow data in a learned fashion; its prediction model assumes that most of the observed deformation can be explained by a rigid transformation (the camera motion), leaving the residual scene flow to small deformations. The solution is massively data-dependent and struggles to generalize to the dual case of an almost static camera viewing tissue undergoing large deformations. SENDD~\cite{schmidt2023sendd} uses a graph neural network to efficiently estimate the depth and scene flow; this framework promises improved generalization by dropping map modeling, but it does not provide camera-motion estimation. The work closest to ours proposed to reconstruct and classify 3D deformable surfaces via multiple overlapping viewpoints~\cite{su2019multicamera}; the method assumes known camera poses and is limited to offline use. To the best of our knowledge, MultiViPerFrOG is the first work to perform real-time simultaneous tracking of camera motion and deformation in a multi-viewpoint setting.

%
\begin{figure*}[t!]
\includegraphics[width=\textwidth]{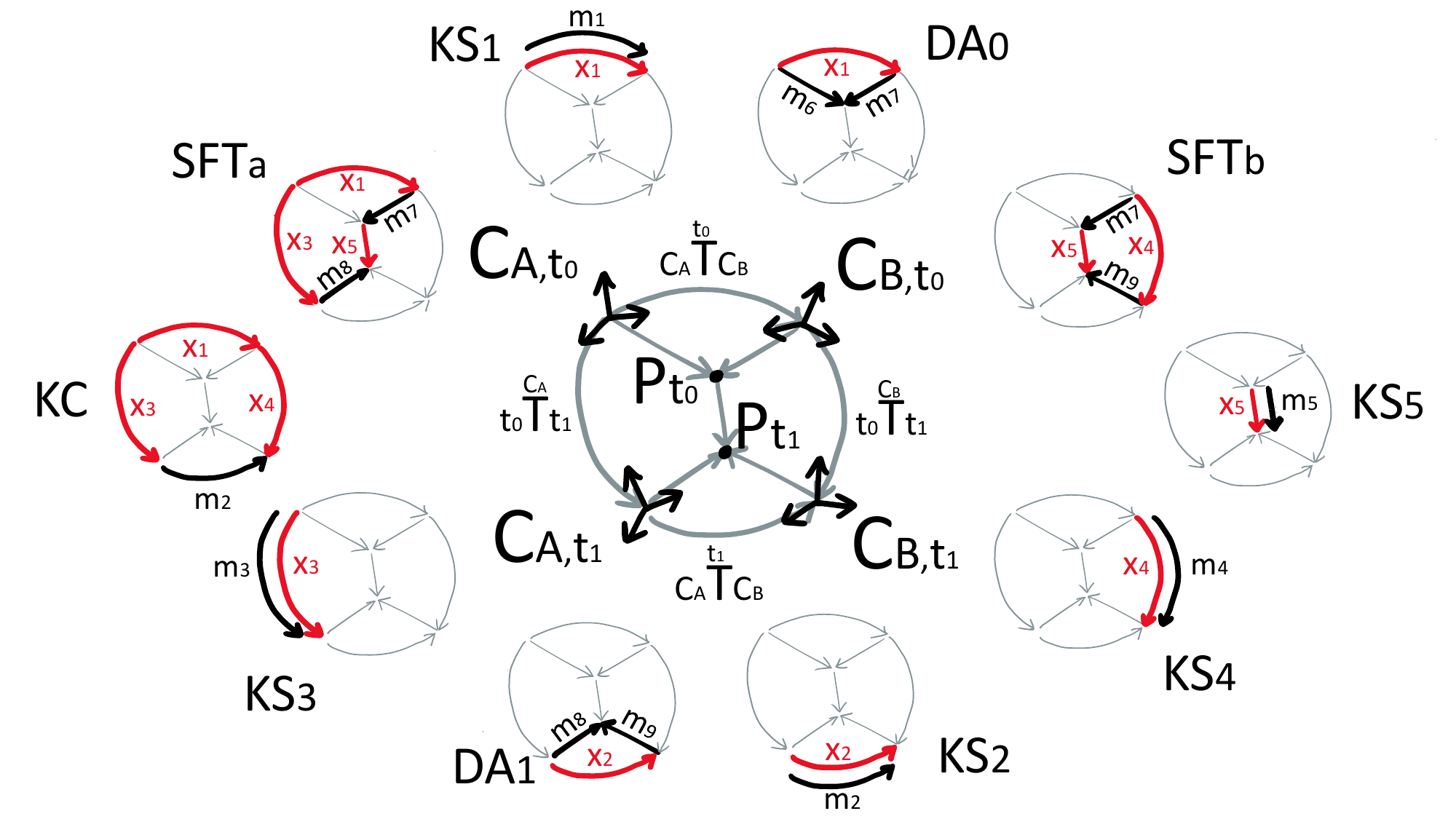}
\caption{Multi-Viewpoint Perception Framework Optimized Globally (MultiViPerFrOG). In the center, a kinematic formalization for two moving cameras $\mathtt{C_{A}}$ and $\mathtt{C_{B}}$ observing a moving point $\mathtt{P}$ between two time instants $\mathtt{t_0}$ and $\mathtt{t_1}$. Around the periphery, the combinations of measures ($m_{1-9}$, black) and parameters ($x_{1-5}$, red) are used to compute the cost functions for each residual block: $DA$ = data association, $SFT$ = scene flow transformation, $KC$ = kinematic chaining, and $KS$ = kinematic supervision.}
\label{framework}
\end{figure*}

\section{Methods}
Let's consider a setting in which multiple depth cameras independently move in space while observing a dynamic and deformable scene. The proposed method \textbf{jointly optimizes the relative transformations between cameras, the ego-motion of each camera, and the absolute scene flow of the visible points in the scene}. Fig.~\ref{framework} shows the kinematic description for two moving cameras $\mathtt{C_{A}}$ and $\mathtt{C_{B}}$ observing a moving point $\mathtt{P}$ between two time instants $\mathtt{t_0}$ and $\mathtt{t_1}$. $\mathtt{P}$ is any point $\mathtt{P_{i}}$ in the scene visible by both cameras at both time instants. Without loss of generality, we present a minimal two-camera formulation that can be extended.
Our model takes up to nine input measures $m_{1-9}$, depending on the experimental setting, to optimize the five output parameters $x_{1-5}$.

\subsubsection*{Input Measures:} Fig.~\ref{framework} shows the spatio-temporal meaning of our measures using black font. First, $m_1$ and $m_2$ are the SE(3) transformations $_\mathtt{C_{A,t}}\mathbf{T}\,_\mathtt{C_{B,t}}$ between the cameras at times $\mathtt{t_0}$ and $\mathtt{t_1}$, respectively. Second, $m_3$ and $m_4$ are the SE(3) ego-motions $_{\mathtt{C_{N,t_0}}}\mathbf{T}\,_\mathtt{C_{N,t_1}}$ for $\mathtt{C_A}$ and $\mathtt{C_B}$, respectively. Third, $m_5$ is a 3D vector describing the absolute motion of the point $\mathtt{P}$ between $\mathtt{t_0}$ and $\mathtt{t_1}$; for convention, we describe $m_5$ in the reference frame of $\mathtt{C_{B,t_0}}$. Fourth, $m_6$ and $m_7$ are the 3D coordinates of $\mathtt{P_{t_0}}$, as measured by $\mathtt{C_{A,t_0}}$ and $\mathtt{C_{B,t_0}}$, respectively. Fifth, $m_8$ and $m_9$ are the 3D coordinates of $\mathtt{P_{t_1}}$, as measured by $\mathtt{C_{A,t_1}}$ and $\mathtt{C_{B,t_1}}$, respectively.

\subsubsection*{Output Parameters:} The optimization outputs one or more estimates $x_{1-5}$, shown in red font in Fig.~\ref{framework}, corresponding to the quantities described by the $m_{1-5}$ measures. The existence of a unique solution for these parameters highly depends on the experimental setting, requiring the availability of all (or a subset) of the measures and also depending on their quality (see Section~\ref{Experiments}). We implemented four types of residual blocks and combined them to estimate $x_{1-5}$:

\textbf{Data Association (DA):} This residual block minimizes the error over the reference frame transformation between two sets of synchronously paired point coordinates. The implemented cost functions are $f_\mathtt{DA_0}(x_1)=m_6 - \, x_1 \cdot\,m_7$ and $f_\mathtt{DA_1}(x_2)=m_8 - \, x_2 \cdot\,m_9$, for measures at times $\mathtt{t_0}$ and $\mathtt{t_1}$, respectively.

\textbf{Scene Flow Transformation (SFT):} 
This residual block minimizes the error between the $\mathtt{P_{t_1}}$ coordinates expressed in a camera frame at $\mathtt{t_0}$ and $\mathtt{t_1}$. It implements the cost functions $f_\mathtt{SFT_A}(x_1, x_3, x_5)=m_8 - \, \mathrm{inv}(x_3) \cdot\, x_1 \cdot\, (m_7 + x_5)$ for $\mathtt{C_{A,t}}$ and $f_\mathtt{SFT_B}(x_4, x_5)=m_9 - \, \mathrm{inv}(x_4) \cdot\, (m_7 + x_5)$ for $\mathtt{C_{B,t}}$.

\textbf{Kinematic Chaining (KC):} This residual block minimizes the error over the concatenation of all the SE(3) transformations across time and space. It implements the cost function $f_\mathtt{KC}(x_1, x_3, x_4)=m_2 - \, \textrm{inv}(x_3) \cdot\, x_1 \cdot\, x_4$.

\textbf{Kinematic Supervision (KS):} This residual block minimizes the error of each parameter against its respective measure. It implements five identical cost functions $f_\mathtt{KS_N}(x_\mathtt{N})=m_\mathtt{N} - \, x_\mathtt{N} $, for $\mathtt{N}\in [1-5]$, one for each $m_\mathtt{N}$, $x_\mathtt{N}$ pair. \\

Considering $\rho$ as the loss function (simple scaled loss) for each residual block and $\mathtt{i}$ as the number of visible points, the resulting minimization problem can be written as:
\begin{equation}
\begin{aligned}
\min_{x} \quad & \frac{1}{2}\Big(\rho_\mathtt{DA_0}\cdot\sum_{\mathtt{i}}{\|f_\mathtt{DA_{0,i}}}\|^2 + \rho_\mathtt{DA_1}\cdot\sum_{\mathtt{i}}{\|f_\mathtt{DA_{1,i}}}\|^2 +\, \rho_\mathtt{SFT_{A}}\cdot\sum_{\mathtt{i}}{\|f_\mathtt{SFT_{A,i}}}\|^2\\ \quad \quad \quad & +\, \rho_\mathtt{SFT_{B}}\cdot\sum_{\mathtt{i}}{\|f_\mathtt{SFT_{B,i}}}\|^2 + \rho_\mathtt{KC}\cdot{\|f_\mathtt{KC}}\|^2 +\, \sum_{\mathtt{N}}\rho_\mathtt{KS_N}\cdot{\|f_\mathtt{KS_N}}\|^2 \Big)
\end{aligned}
\label{DefSF}
\end{equation}
We implemented our optimization routines in \texttt{C++} using \texttt{CERES}~\cite{Agarwal_Ceres_Solver_2022}, an open-source large-scale non-linear optimization library that showed the best solving performance on the NIST problems~\cite{Mondragon05-JMASM-Comparison}. We chose this specific solver due to its robustness, flexibility in the choice of the factorization, and compatibility with \texttt{Eigen} for defining our linear algebra. In particular, we used the \texttt{EigenQuaternion\-Manifold} component to perform Lie algebra operations and the \texttt{AutoDiffCost\-Function} for automatic differentiation. The reported experiments used sparse Cholesky factorization to prioritize real-time performance. While higher parallelization can be achieved in \texttt{CERES} via \texttt{CUDA} or \texttt{OPENMP}, we report (Fig.~\ref{Fig:experiment3}) computation times for a single thread on an i9 3\,GHz CPU to highlight the portability of our framework. 


\begin{figure*}[tb]
\includegraphics[width=\textwidth]{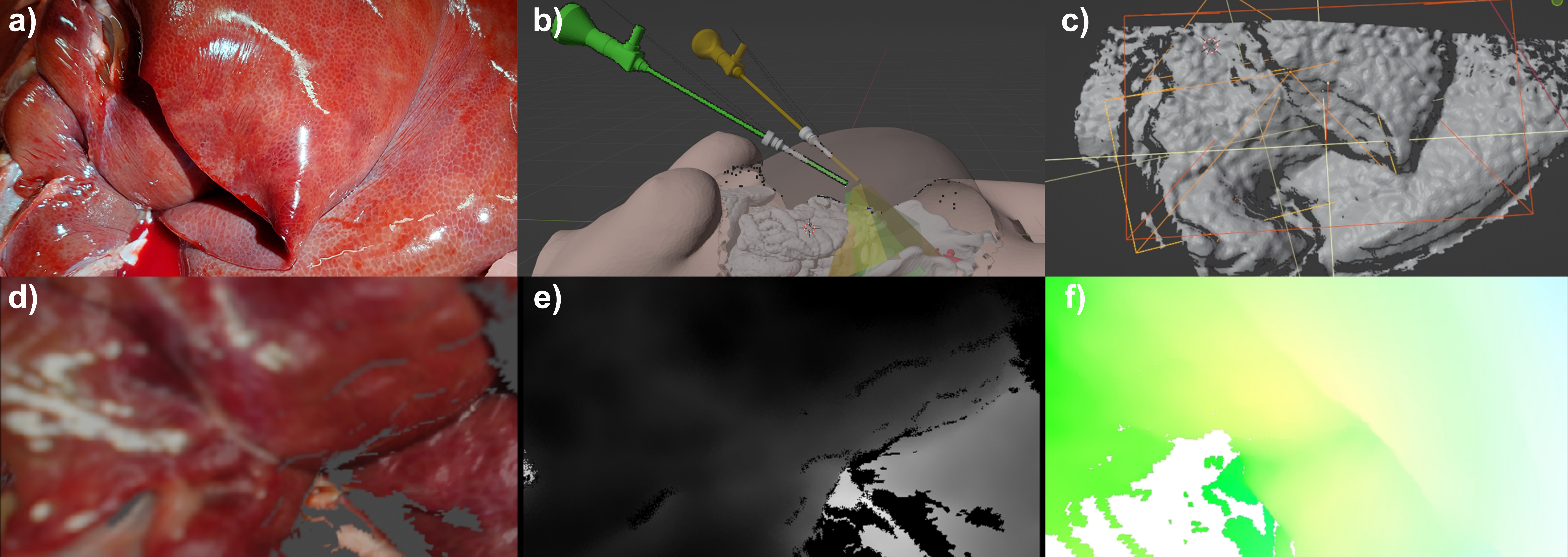}
\caption{Workflow for synthetic dataset generation. a) Ex vivo porcine liver captured for organ mesh. b) Simulated laparoscopic scene with two cameras. c) Liver mesh. d) Sample RGB, e) depth, and f) optical flow outputs from one virtual depth camera.}
\label{dataset}
\end{figure*}
\begin{figure*}[tb]
\includegraphics[width=\textwidth]{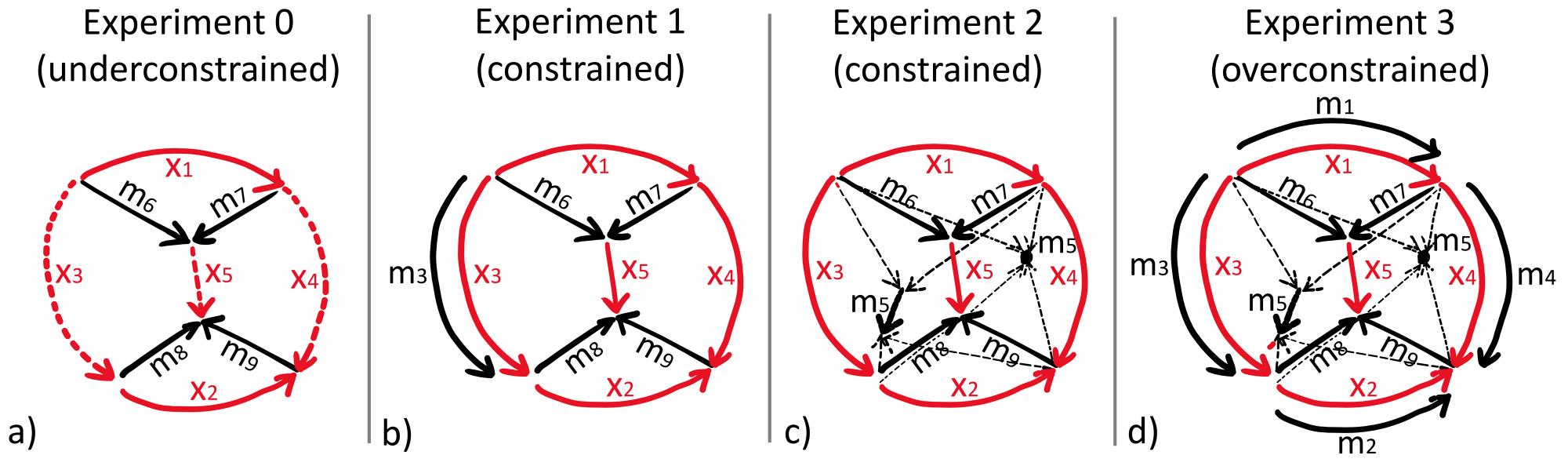}
\caption{Experiment overview. a) Experiment 0: The optimization is underconstrained as infinitely many combinations of the unknown parameters ($x_{3-5}$) can explain the relevant measures ($m_{6-9}$). $m_{1-2}$ do not remove this ambiguity. b) Experiment 1: Measuring the odometry of one camera ($m_3$) constrains the problem to a unique solution. This setting is valid for the camera being either static or freely moving. c) Experiment 2: Measuring a number of absolute scene-flow values ($m_5$) also constrains the problem to a unique solution. These measures can be either static or moving points. d) Experiment 3: All the measures $m_{1-9}$ are available and overconstrain the problem.}
\label{Fig:experiments}
\end{figure*}

\section{Experiments}
\label{Experiments}

\textbf{Dataset Generation:} While accurately measuring kinematic transformations in a lab setting is quite straightforward, there is currently no simple standardized method to obtain ground-truth scene flow data from a real deforming surface. For this reason, the computer-vision community has widely adopted the use of synthetic datasets for benchmarking optical~\cite{Butler:ECCV:2012} and 3D flow~\cite{mehl2023spring}. Similarly, VisionBlender was proposed for surgical ground-truth dataset generation~\cite{cartucho2020visionblender}. Its Blender plugin was recently adapted to include deformation while generating depth and optical flow to supervise deformable odometry estimation in endoscopy~\cite{recasens2023drunkard}. We built on these works and integrated the patch of~\cite{mehl2023spring} to allow VisionBlender to output 3D deformation fields as seen by the moving camera (relative scene flow) in addition to RGB, depth, and camera poses. 

We imported the mesh of a liver captured via lidar from an ex-vivo porcine sample (Fig.~\ref{dataset}a) and reproduced a laparoscopic scene in Blender. As in the minimal setting described above, two cameras rotate and translate independently while observing a deforming anatomical surface (liver) with partial view-field overlap (Fig~\ref{dataset}b).  
The two point clouds obtained from the depth maps are registered together using the ground-truth kinematic transformations ($m_{1-4}$). A kNN search ($r=0.00005$~m) is performed on the two clouds to find a set of matching points $\mathtt{P_{i,t_0}}$ ($m_6$, $m_7$) in the overlapped region. The relative scene-flow data are added to those points to produce the point coordinates $\mathtt{P_{i,t_1}}$ ($m_8$, $m_9$) in the camera frames at time $\mathtt{t_1}$. A graphical representation of this vector addition is available in the supplementary material. Finally, relative scene flow vectors are transformed in the camera frame at time $\mathtt{t_0}$ to get the absolute scene flow ($m_5$).

\textbf{Metrics:} To quantify the optimization results against ground-truth data, we adopted the average distance (ADD)~\cite{Hinterstoisser13-ACCV-Pose,xiang2018posecnn}; lower values of ADD indicate better performance. We chose this scene-space metric as it can be used equivalently for both SE(3) and scene-flow data. For the kinematic outputs ($x_{1-4}$), $\mathtt{ADD_{outTf}}$ is computed as the average by-index distance between a set of points transformed by the ground-truth SE(3) and the estimated one.
For the absolute scene-flow output ($x_5$), $\mathtt{ADD_{outSF}}$ simplifies to the average by-index distance between the ground-truth and estimated vector fields. When adding noise to our input measures, we used an equivalent set of metrics ($\mathtt{ADD_{inDA}}$, $\mathtt{ADD_{inTf}}$, $\mathtt{ADD_{inSF}}$), deployed to quantify the distance of the noisy inputs from the ground-truth ones. Gaussian noise is added to 3D vectors and translations in the Euclidean space and added to rotations in the quaternion manifold space.\looseness-1

\textbf{Experiment 0 (underconstrained):} When no ego-motion ($m_{3-4}$) information is available, nor any knowledge of the absolute point motions in the scene ($m_{5}$), the optimization lacks sufficient constraints. The data association between measured point coordinates ($m_{6-9}$) is sufficient to estimate the between-camera transformations ($x_{1-2}$), but there are infinitely many combinations (Fig.~\ref{Fig:experiments}a) of camera ego-motions ($x_{3-4}$) and absolute point motions ($x_{5}$) that, given the point coordinates at $\mathtt{t_0}$ ($m_{6-7}$), would explain the point coordinates at $\mathtt{t_1}$ ($m_{8-9}$) computed with relative scene-flow data. Measures of the between-camera transformations ($m_{1-2}$) add constraints but do not resolve this ambiguity.

\begin{figure*}[tp]
\includegraphics[width=\textwidth]{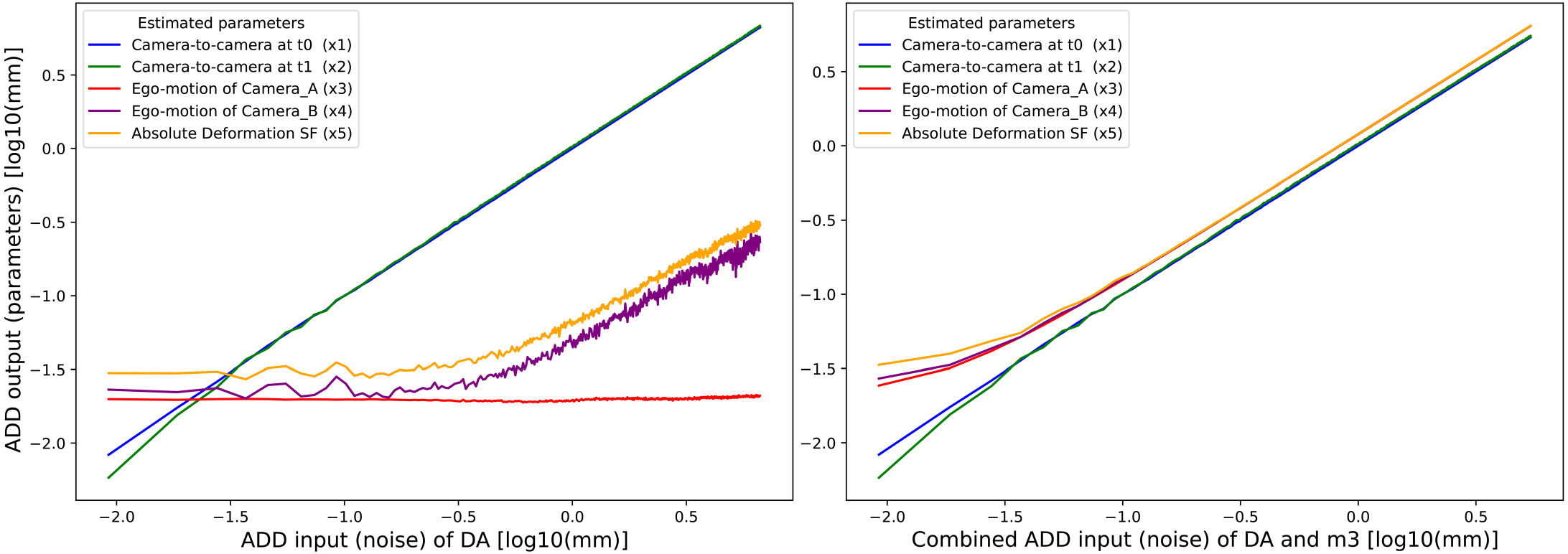}
\caption{Experiment 1: Left) Increasing [0--10~mm] noise is added to the data association (DA) between the two cameras. The ego-motion of one camera is exactly known ($m_3$), whether static or moving. Right) The same noise is added to both DA and $m_3$.}
\label{Fig:experiment1}
\end{figure*}

\textbf{Experiment 1 (kinematic prior):} Consider a case in which one camera's ego-motion is known, either static or moving (Fig.~\ref{Fig:experiments}b). This example is relevant to real laparoscopic surgery where one camera may be attached to a rigid frame or robotic arm while another camera is handheld. Similarly, it extends to multi-camera robot-assisted surgery where one viewpoint is kept static during tool-tissue interaction, while others are handheld or controlled by a second agent (human or algorithm). 
The left panel of Fig.~\ref{Fig:experiment1} shows the effect of increasing noise in the data association when the exact ego-motion of one camera is available. The between-camera transformations ($x_{1-2}$) are linearly affected by the noise injected via $\mathtt{ADD_{inDA}}$ in the point coordinates at $\mathtt{t_0}$ ($m_{6-7}$). The known ego-motion ($x_{3}$) is not affected, and the remaining two parameters ($x_{4-5}$) register sub-millimeter accuracy consistently when $\mathtt{ADD_{inDA}}$ ranges from 0 to 10~mm. The right panel of Fig.~\ref{Fig:experiment1} shows how all the output parameters are linearly affected by the combined application of $\mathtt{ADD_{inDA}}$ (to $m_{6-7}$), and $\mathtt{ADD_{inTf}}$ (to $m_3$).

\begin{figure*}[tp]
\includegraphics[width=\textwidth]{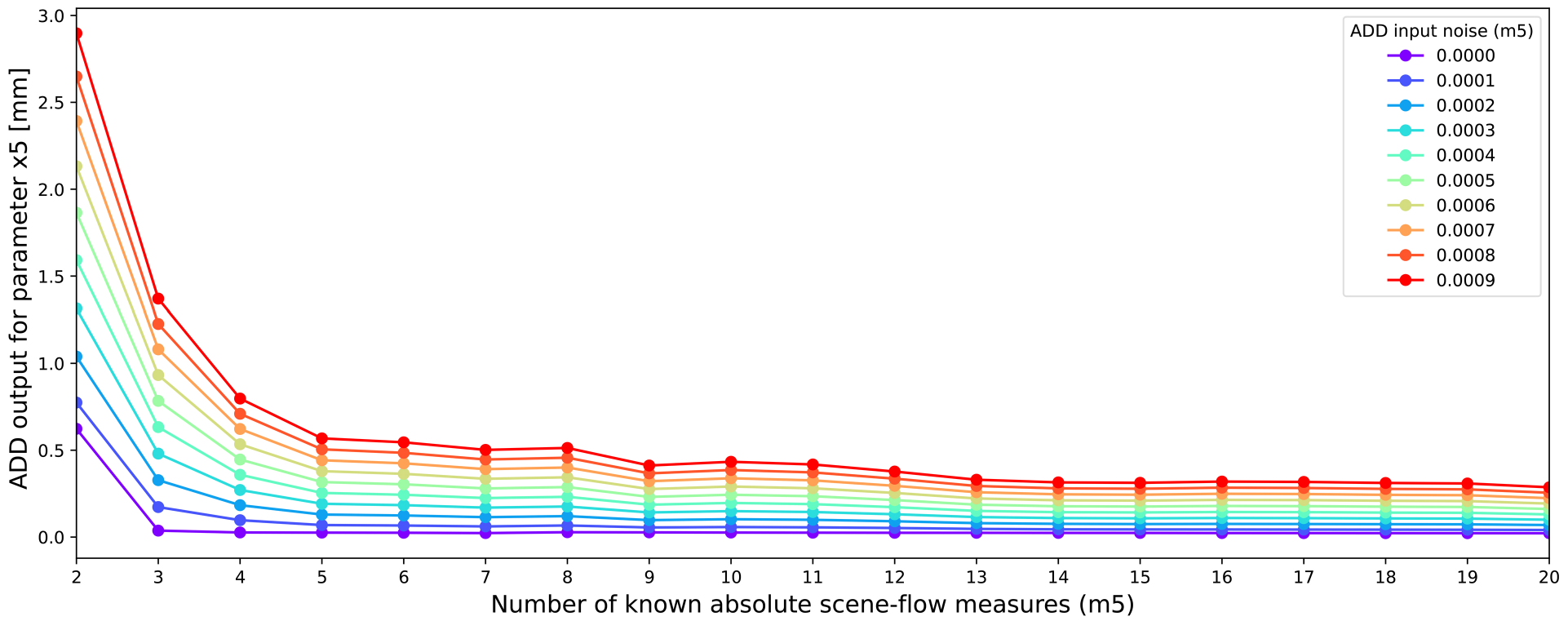}
\caption{Experiment 2: The number of known absolute scene-flow measures ($m_5$) is increased, reducing the error of the respective parameter ($x_5$). Increasing [0--1~mm] noise is added to $m_5$. No measure of the ego-motion of the cameras ($x_{3-4}$) is used.}
\label{Fig:experiment2}
\end{figure*}

\textbf{Experiment 2 (scene-flow prior):} In this experiment, no measure of the camera ego-motions ($m_{3}$ or $m_{4}$) is introduced, while we assume partial knowledge of the absolute scene flow ($m_{5}$) to be available (Fig.~\ref{Fig:experiments}c). This example is relevant to any surgical setting in which some points or areas in the scene can be labeled (manually or algorithmically) as static. Similarly, it extends to robot-assisted surgery in which some elements of the scene (e.g., the surgical instruments) can be tracked in an absolute reference frame and therefore used by the optimization to infer the ego-motions of the cameras. Fig.~(\ref{Fig:experiment2}) shows the combined effect on the scene-flow output metric ($\mathtt{ADD_{outSF}}$) of the number of measured absolute scene-flow vectors ($m_5$) and the level of noise applied to them. The minimum number of points required to constrain the system is two, given that their trajectories between $\mathtt{t_0}$ and $\mathtt{t_1}$ are not colinear.

\begin{figure*}[tp]
\includegraphics[width=\textwidth]{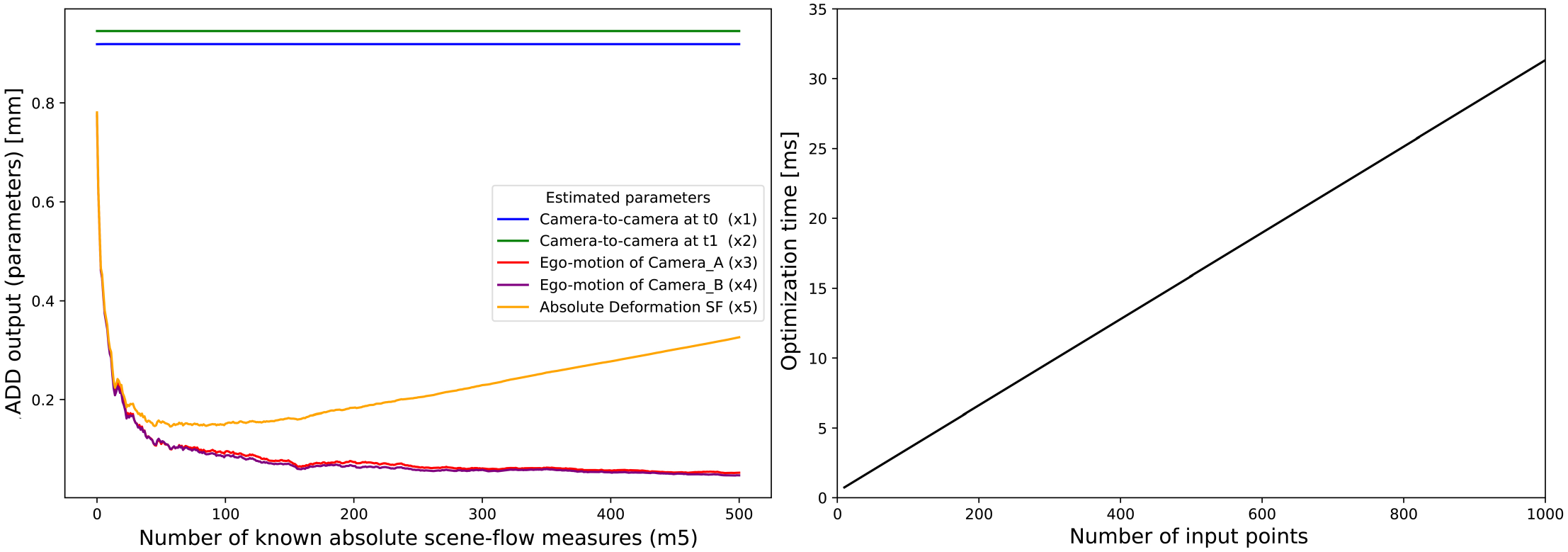}
\caption{Experiment 3: Constant (1~mm) noise is independently added to all the measures ($m_{1-9}$). Left) An increasing number of noisy scene-flow measures in $m_5$ impacts only three of the output parameters: $x_3$, $x_4$, and $x_5$. Right) The optimization time increases linearly with the number of input points (five values known in $m_5$).}
\label{Fig:experiment3}
\end{figure*}

\textbf{Experiment 3 (combined priors):} Finally, we report on the performance of our optimization method when all the measures ($m_{1-9}$) are available (Fig.~\ref{Fig:experiments}d). We added, simultaneously and independently, noise on the order of magnitude of 1~mm to all measures ($\mathtt{ADD_{inTf}}$ to $m_{1-4}$, $\mathtt{ADD_{inSF}}$ to $m_5$, and $\mathtt{ADD_{inDA}}$ to $m_{6-9}$). Note that the scene flows measured in the simulation are in the 1--5~mm range. The left panel of Fig.~\ref{Fig:experiment3} shows the impact of the number of points provided to the algorithm in $m_5$, with the error stabilizing around 20. The right panel shows the total convergence time required for this experiment as a function of the number of input points. Our optimization can process 500 points in about 15~ms.

While a full characterization of MultiViPerFrOG using real surgical data is needed, it is out of the scope of this manuscript. With the reported experiments, we highlight the flexibility of our approach in handling different setups and noisy sources of information. More advanced prior knowledge (e.g., camera-motion mechanical constraints, vector-field smoothing) could easily be integrated into our system by modeling additional residual blocks that can restrict the convergence basin in particular ways.

    
\section{Conclusion}
We presented MultiViPerFrOG, a global optimization framework that builds on top of low-level surgical perception modules (data association, depth, and relative scene flow) to simultaneously estimate multiple camera motions and absolute tissue deformation. Our framework is learning-free, easy to customize, and optimized for real-time performance via robust software libraries. We believe such an approach is a fundamental integration block that will allow the development of the computer-assisted-surgery technologies of the future. For example, the presented method could be used to allow a surgeon to see the surgical field from another viewpoint that moves in real time, or to enable partially autonomous surgical tool movements relative to the constantly deforming soft tissue of the patient's anatomy. 
Our code and data will be released upon paper acceptance. 


\begin{credits}
\subsubsection*{Acknowledgments} 
The authors thank the International Max Planck Research School for Intelligent Systems (IMPRS-IS) for supporting Guido Caccianiga and the Max Planck ETH Center for Learning Systems (CLS) for supporting Guido Caccianiga and Julian Nubert.

\end{credits}

\bibliographystyle{splncs04}
\bibliography{main}

\appendix

%

%

%

%
%

%
\title{Supplementary Material for \\MultiViPerFrOG: A Globally Optimized Multi-Viewpoint Perception Framework for Camera Motion and Tissue Deformation}
\author{}
\institute{}
\maketitle              

\begin{figure}
\includegraphics[width=\textwidth]{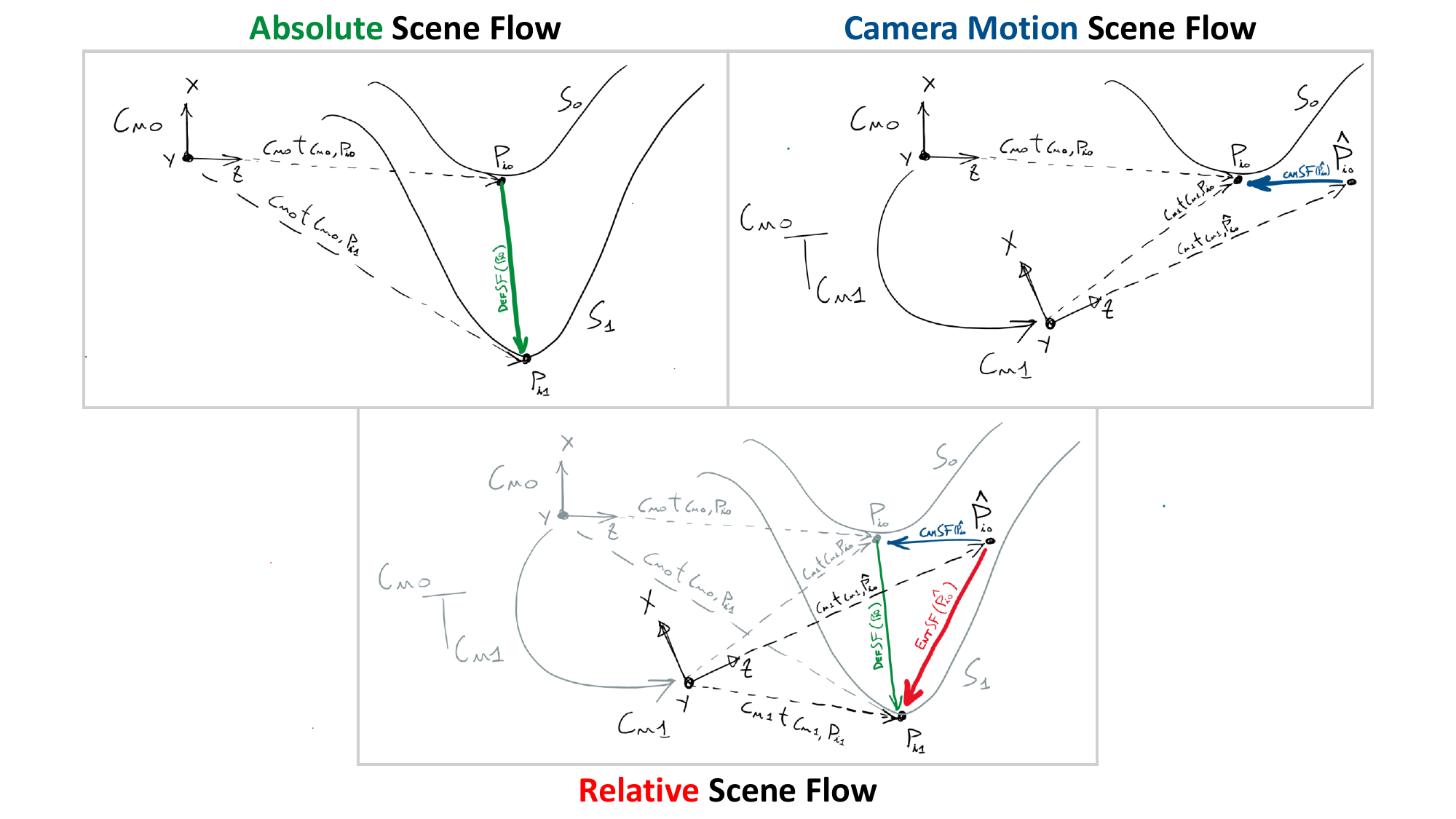}
\caption{Kinematic representations (shown in the $X$-$Z$ plane) of the possible relative motions and resulting scene flows (colored arrows) between a camera $\mathtt{C}$ and a point $\mathtt{P}$ in its view field at two time instants $\mathtt{t_0}$ and $\mathtt{t_1}$. a) Absolute scene flow: a static camera observes a moving point (green arrow). b) Camera scene flow: a moving camera observes a static point (blue arrow). The point $\mathtt{\hat{P}_{t_o}}$ represents the coordinates of the point $\mathtt{P_{t_o}}$, as measured by $\mathtt{C_{t_o}}$, applied in the reference frame of $\mathtt{C_{t_1}}$. c) Relative scene flow: a moving camera observes a moving point (red arrow).} \label{fig1}
\end{figure}


\end{document}